\documentclass[11pt,letterpaper]{article}
\usepackage{emnlp2017}
\usepackage{times}
\usepackage{latexsym}
\usepackage{multirow}
\usepackage{graphicx}
\usepackage{url}

\emnlpfinalcopy

\def\emnlppaperid{***}

\title{Predicting Specificity in Classroom Discussion}

\author{Luca Lugini \and Diane Litman\\
  Computer Science Department \& Learning Research and Development Center\\
  University of Pittsburgh \\
  Pittsburgh, PA 15260 \\
  {\tt \{lucalugini,litman\}@cs.pitt.edu}}

\date{}

\begin{document}

\maketitle

\author{Luca Lugini \and Diane Litman\\
  Computer Science Department \& Learning Research Development Center\\
  University of Pittsburgh \\
  Pittsburgh, PA 15260 \\
  {\tt \{lucalugini,litman\}@cs.pitt.edu}}

\begin{abstract}
High quality classroom discussion is important to student development, 
enhancing abilities to express claims, reason about other students' claims, and retain information for longer periods of time.
Previous small-scale studies have shown that one indicator of classroom discussion quality is specificity.
In this paper we tackle the problem of predicting specificity for classroom discussions. We propose several methods and feature sets capable of outperforming the state of the art in specificity prediction. Additionally, we provide a set of meaningful, interpretable features that can be used to analyze classroom discussions at a pedagogical level.
\end{abstract}

\section{Introduction}
Classroom discussion plays an important role in the learning process.
It has been shown that reasoning, reading, and writing skills can be positively affected by high-quality student-centered classroom discussion in the context of English Language Arts (ELA) classrooms \cite{Reznitskaya:13,Graham:07,Applebee:03}.
High quality discussions encourage student-to-student talk, negotiation of claims, supporting claims with evidence, and reasoning about those claims.
Although the effectiveness of particular kinds of claims, evidence and reasoning can vary across disciplines, Chisholm and Godley \shortcite{Chisholm:11} and Lee \shortcite{Lee:06} showed that the specificity of these argument moves is related to discussion quality.
These findings are  based on a largely qualitative analysis of a single classroom discussion
that relied on the manual annotation of specificity and discussion quality.
The proposed method  in this paper will help address this limitation by making the annotation of specificity automatic.

Specificity is defined by the Oxford Dictionary as ``The quality of belonging or relating uniquely to a particular subject" \footnote{https://en.oxforddictionaries.com/definition/specificity}.
Natural language processing (NLP) techniques can be used to facilitate the analysis of classroom discussion and of specificity. Chen et al. \shortcite{Chen:14} developed a tool for teacher self-assessment of classroom discussion through the analysis of the frequency of participation of students in the discussion, and teacher-student turn patterns. Blanchard et al. \shortcite{Blanchard:16} developed a system for detecting teacher questions from classroom discussion recordings. These works, however, do not take into account the actual student discussion content. 
Speciteller \cite{Li:15} is a current state of the art method for predicting sentence specificity. It was developed by analyzing newspaper articles to distinguish between general and specific sentences.
Spoken and written language differ in grammatical structure, contextual influence, and cognitive process and skills  \cite{Chafe:87,Biber:88}. As such we believe that using Speciteller as-is on classroom discussions will lead to sub-optimal performance, which we can improve.

In this paper we propose a method to automatically determine specificity of student turns at talk in high school ELA classroom discussions of texts. The contributions of this paper are twofold.
For the educational community this work will enable the exploration of hypotheses concerning specificity and discussion quality over large datasets, spanning multiple classes and including a large number of students, which would otherwise require a prohibitive amount of work for manually annotating data.
Additionally, we develop a set of pedagogically meaningful features which can be used to understand important elements of highly specific discussions.
For the NLP community, we make the following contributions: we experimentally evaluate the performance of prior approaches for predicting specificity in a new domain; we compare between different feature sets and algorithms; finally, we provide a model for predicting specificity tailored to spoken dialogue and in an educational context, which outperforms the current state of the art.

\section{Related Work}
To the best of our knowledge, this is the first work to analyze specificity of spoken dialogue, and more precisely in classroom discussions.
Louis and Nenkova \shortcite{Louis:11} analyzed specificity in news articles and their  summarizations. Their proposed method leverages a combination of lexical and syntactic features and annotated data from the Penn Discourse Treebank  to train a logistic regression classifier. They used the trained model to analyze differences in specificity between human-written and automatically-generated summaries of news articles.
Li and Nenkova \shortcite{Li:15} developed Speciteller, a tool for predicting the specificity score of sentences. Specificity was defined in relation to the amount of details in a sentence. This tool uses a set of shallow features (described in Section \ref{sec:speciteller_features}) and two dense word vector representations to train two logistic regression models on  Wall Street Journal articles. Additionally, they improved classification accuracy by using a semi-supervised co-training method on over thirty thousand sentences from the Associated Press, New York Times, and Wall Street Journal. Finally, Li et al. \shortcite{Li:16} 
improved the  annotation scheme used in \cite{Louis:11,Li:15} by considering contextual information, 
and by using a scale from 0 to 6 rather than binary specificity annotations.
Our annotation scheme is based on prior educational work in coding specificity~\cite{Chisholm:11}, and our prediction models will incorporate features used by Speciteller.

Like other machine learning-based methods, Speciteller is highly dependent on its training data. Since our objective is to analyze classroom discussion, we also draw on work that has used Speciteller to analyze data that is more similar to our corpus.
Swanson et al. \shortcite{Swanson:15} analyzed online forum dialogues in the context of argument mining. By performing feature selection they observed that argument quality is highly correlated with specificity as measured by Speciteller across multiple topics. We believe there might be a correlation between specificity and  other features used in their work (described in Section \ref{sec:literature_features}) to predict argument quality, therefore we used some of these features in our approach.

\section{Dataset Description}
The dataset for this work consists of manually transcribed text-based classroom discussions from  English Language Arts high school  classes. Text-based discussions are about a ``text'' (e.g., literature such as {\it Macbeth} and {\it Memoir of a Geisha}, a news article, a speech, etc.) and can either be mediated by a teacher or conducted exclusively among students. The number of students per discussion ranges from 5 to 13 in our dataset. 

Motivated by  Chisholm and Godley's ~\shortcite{Chisholm:11} and Lee's \shortcite{Lee:06} coding of classroom discussions, a codebook for manually annotating student argument moves and specificity has been developed.
Each student turn at talk is labeled for: $(i)$ specificity (low, medium, high); $(ii)$ argument move (claim, evidence, warrant).
Specificity was labeled at the level of argument move: each transcript was preprocessed by one of the annotators and a decision was made on whether to segment each turn at talk into multiple ones if the turn at talk could potentially contain multiple argument move types.
The following aspects were considered when labeling specificity for a turn at talk:
\begin{enumerate}
\item it involves one character or scene;
\item it gives substantial qualifications or elaboration;
\item it uses content-specific vocabulary;
\item it provides a chain of reasons. 
\end{enumerate}
If none of the four elements was present, or if the turn at talk refers to all humans or the text in general, the turn at talk is labeled as low specificity. Medium specificity turns at talk contain one of the four elements, while high specificity ones contain at least two of the four elements.

Table \ref{tab:examples} shows examples of specificity annotation from one of the discussions in our dataset about the book {\it Death of a Salesman}.
\begin{table*}[t]
\centering
\begin{tabular}{|p{0.81\linewidth}|c|}
\hline
\textbf{Turn at talk} & \textbf{Specificity} \\ \hline
It's just kind of a maintaining personality            & low         \\ \hline
Yeah because she just couldn't- I mean, it's not a fake personality, but it's kind of like superficial            & med         \\ \hline
At one point, I don't even think she's concerned that like with her sons as much as she is with Willy, or you know, she's just focusing most of her attention and comfort on Willy and um, when Biff and Happy are there it makes him, like, [inaudible]. I think she's trying to like, you know, be the bridge between them and Willy.             & high        \\ \hline
\end{tabular}
\caption{Examples of turns at talk for different specificity classes.}
\label{tab:examples}
\end{table*}
The first turn at talk in the table was labeled as low specificity because the claim made by the student was unsubstantiated. The student did not give a definition of what maintaining personality means in this context, nor did they mention the reasons for making such a claim.
The second turn at talk in the table, although not providing considerable elaboration, is clearly about one individual character in the book. As such, it is classified as medium specificity.
The third turn at talk is classified as high specificity because the statement is particular to one or a few selected characters, and the student shows a clear chain of reasoning.

The dataset spans $23$ classroom discussions and over $2000$ turns at talk. 
Two pairs of annotators coded specificity for 5 and 9 transcripts respectively, while the remaining 9 transcripts were single-coded.
Inter-rater reliability  on specificity labels for the two annotator pairs as measured by quadratic-weighted Cohen's Kappa is $0.714$ and $0.9$, indicating substantial agreement and almost perfect agreement, respectively.\footnote{Although argument move types are not used in our study, 
Kappa for the two annotator pairs were $0.75$ and $0.89$.}
A gold standard set of labels for each double-coded discussion was obtained by resolving the disagreements between the two annotators.
Table \ref{data_statistics} shows the distribution of specificity classes in our dataset.
\begin{table}[h]
\centering
\begin{tabular}{|p{3cm}|c|c|c|}
\hline
\multirow{2}{*}{\textbf{Turns at talk}} & \multicolumn{3}{c|}{\textbf{Specificity}}      \\ \cline{2-4} 
                                                  & \textbf{Low} & \textbf{Medium} & \textbf{High} \\ \hline
2057                                               & 730           & 974             & 353            \\ \hline
\end{tabular}
\caption{Dataset statistics.}
\label{data_statistics}
\end{table}

\section{Proposed Method}
This section provides a description of Speciteller \cite{Li:15} and additional features and models that we propose to predict specificity.

\subsection{Speciteller tool}
The baseline for testing our hypotheses consists of using Speciteller out of the box to predict the specificity of each turn at talk. Speciteller accepts a string as input and outputs a specificity score in the range $[0,1]$, where 0 indicates general sentences and 1 indicates specific sentences. Since the unit of analysis for the current work is a turn at talk, which may consist of multiple sentences, we evaluated the performance of Speciteller in several scenarios (e.g. sentence, turn at talk). We found that the best results are obtained when using the complete turn at talk as input to Speciteller.
In order to convert the numeric specificity score into a specificity class (i.e. low, medium, or high) we set two thresholds $t_1$ and $t_2$, then labeled turns at talk with specificity score $s \leq t_1$ as low, those with score $t_1 < s \leq t_2$ as medium, and those with score $s > t_2$ as high.
The optimal thresholds were found by starting at $0$ and iteratively increasing them by $0.001$ at each step, while saving the best results.
The values for the optimal thresholds are: $t_1 = 0.02$ and $t_2 = 0.78$.
It is important to note that this represents the upper bound for Speciteller's performance. Finding the optimal thresholds is not trivial and in practice it could be done through cross-validation. 

\subsection{Speciteller feature set}
\label{sec:speciteller_features}
The initial set of features we evaluated was that used in Speciteller. We extracted features from each turn at talk using the source code provided by Speciteller\footnote{ https://www.cis.upenn.edu/~nlp/software/speciteller.html}. In their proposed method, Li and Nenkova extracted two categories of features, a shallow feature set and a word embeddings set, and used them for two separate classifiers. In this work, we concatenate both shallow features and word embeddings to form a single feature vector. We will refer to these features as the Speciteller set.
Shallow features for each sentence consist of: number of connectives, sentence length (number of words), number of numbers, number of capital letters, number of symbols (including punctuation), average number of characters for the words in the sentence, number of stopwords (normalized by sentence length), number of strongly subjective and polar words (using the MPQA \cite{Wilson:09} and the General Inquirer \cite{Stone:63} dictionaries), average word familiarity and imageability (using the MRC Psycholinguistic Database \cite{Wilson:88}), average, maximum, minimum inverse document frequency values.
Word embeddings features consist of the average of 100-dimensional vectors for each word in the sentence. The embeddings were provided by Turian et al. \shortcite{Turian:10} and trained on a corpus consisting of news articles.

\subsection{Online dialogue features}
\label{sec:literature_features}
While extracting arguments from online forum dialogues, Swanson at al. \shortcite{Swanson:15} found that Speciteller scores (as a measure of specificity) are highly correlated with argument quality. In addition to Speciteller scores, their model used several feature sets. While not explicitly stated by the authors, we believe there might exist a correlation between specificity and the other feature sets.
We will add the following sets of features to the features already present in Speciteller.\\
{\bf Semantic features}\footnote{The name of the feature set in the original paper is semantic-density features; we use semantic features for brevity.} \quad The number of pronouns present in a given turn at talk. Descriptive statistics for word lengths: minimum, maximum, average, and median length of the words in a turn at talk. It is worth noting that the average word length differs from the one implemented in Speciteller as this feature keeps punctuation into account. Number of occurrences of words of length 1 to 20: one feature for each word length - words longer than 20 characters will be counted in the feature for length 20.\\
{\bf Lexical features} \quad N-gram language models are often powerful features, but one drawback is their dependence on specific domains. Since we plan to build a model for predicting specificity which is able to generalize to multiple topics, we did not use the raw N-gram features. To alleviate this problem, we used the term frequency - inverse document frequency (tf-idf) feature for each unigram and bigram in the corpus with frequency of at least 5. Descriptive statistics of lexical features for each turn at talk, namely minimum, maximum, and average, were also used.\\
{\bf Syntactic features} \quad To mitigate the data sparsity  that impacts word n-grams, and to get more generalizable features, we extracted unigrams, bigrams, and trigrams of Parts Of Speech (POS) tags, using the Natural Language Toolkit \cite{Bird:09}.

\subsection{Additional feature sets}
\label{sec:additional_features}
In addition to the previous feature sets, we also extracted the following feature sets which we believe are able to capture specificity with respect to the educational domain of ELA text-based classroom discussions.\\
{\bf Pronoun features} \quad Pronouns are grammatical units that might help us gain useful information about the focus of a turn at talk. For example, if the pronoun ``she" is present in a turn at talk, the student might likely  be referring to one specific character, which is one of the aspects considered when annotating specificity. Therefore we extracted a set of the following pronoun features: binary feature indicating presence/absence of pronouns; total number of pronouns in the turn at talk\footnote{This feature differs from that described in section \ref{sec:literature_features}: the feature from the online dialogue set only considers deictic pronouns.}; the numbers of first, second, and third person pronouns; the number of singular and plural pronouns; the number of pronouns for each of the following categories: personal, possessive, reflexive, reciprocal, relative, demonstrative, interrogative, indefinite.\\
{\bf Named entities} \quad Named entities might give us a sense of characters or places that students discuss, with respect to specificity. For example, saying ``I did not like Biff" is more specific than saying ``I did not like one of the characters" as it points out which of the characters a student might not like. For this task we used the Stanford Named Entity Recognizer \cite{Finkel:05} (NER) with the pre-trained 3 class model detecting location, person and organization entities. We extracted the following features: a binary feature indicating the presence/absence of any named entity; a binary feature indicating presence/absence of each of the three named entity classes; the total number of named entities; the total number of named entities per class. We complemented the previous counts by adding a normalized feature, with respect to the length of the turn at talk, for each of them.\\
{\bf Book features} \quad Since our dataset consists of text-based discussions, we might be able to leverage information about the texts (i.e. books) for each discussion to understand how much each turn at talk is related to the book or its characters. First, a manually-created summary and a list of characters for each book were obtained from the web, using Wikipedia when possible or Sparknotes as an alternative. Then, the following character-related features were extracted from each turn at talk: a binary feature indicating the presence/absence of a character's name; the number of characters mentioned; the number of characters mentioned normalized by the length of the turn at talk. A character was counted by matching each word in the turn at talk to their first name, last name, or their nickname.
Additionally the following summary related features were extracted: the number of overlapping words with the turn at talk; Jaccard similarity between the turn at talk and the summary; tf-idf based cosine similarity between the summary and the turn at talk. We extracted the summary related features in two different settings: considering the book summary as a single entity; computing the similarity between the turn at talk and each sentence in the summary, then picking the maximum.
All features were extracted after removing stopwords from the turn and summary.\\
{\bf Embeddings} \quad Li and Nenkova \shortcite{Li:15} used sentence embeddings based on word embeddings in order to increase the accuracy of Speciteller. The sentence embeddings were obtained by computing the average of pre-trained word embeddings for each word in the sentence. We believe our method can further benefit from sentence embeddings specifically trained on our corpus and optimized for our target: predicting specificity. We generated embeddings by training a character-level Long-Short Term Memory (LSTM) network \cite{Hochreiter:97}, using it as an encoder on the turns at talk from our corpus.
Each turn at talk, which might consist of multiple sentences, represents one sequence (training sample) for the LSTM training. 
Since punctuation is not very meaningful given that we are analyzing spoken discussions, all characters that are not letters or numbers are ignored. Inputs for the LSTM consist of one-hot (1 X N) encoding of individual characters.

\begin{figure}[t]
\begin{center}
  \includegraphics[scale=0.35]{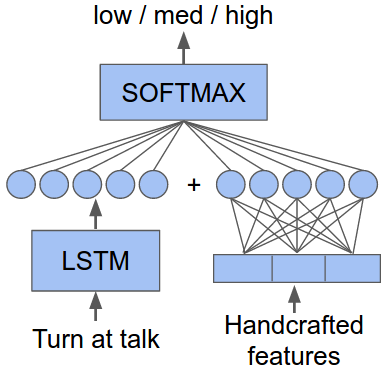}
  \end{center}
  \caption{Network setup for training neural network-based embeddings.}
  \label{fig:lstm}
\end{figure}

The neural network is trained by using the hidden state of the LSTM unit at the end of the turn at talk as embedding, feeding it to a softmax classifier for predicting specificity, and back-propagating errors. Cross-entropy was used as the objective function to optimize during training. A disadvantage of neural network models is the fact that their large number of parameters requires extensive amount of data to show their expressive power. Given the size of our training data we try to mitigate this problem by merging the embeddings for a turn at talk with handcrafted features. Ideally we would combine embeddings with all the features described previously but the resulting model would be far too large for our dataset, therefore we chose to use the $Speciteller + Semantic$ feature set for this task.
The training procedure changes slightly: a turn at talk is propagated through the LSTM resulting in a fixed size embedding; handcrafted features are extracted from the turn at talk, concatenated to form a vector, and a fully-connected layer is applied to those; the output of the fully-connected layer is concatenated with the embedding, and given as input to a softmax classifier to predict specificity. A graphical overview of the model is given in Figure \ref{fig:lstm}.

It is important to note that the neural network for embeddings and the classifier are jointly trained, therefore the embeddings are specifically tailored to encode information regarding specificity.
The Keras library \cite{Chollet:15} was used for extracting sentence embeddings as well as for evaluating performance of the softmax classifier.

{\bf Pedagogical feature set} \quad In addition to maximizing kappa for specificity prediction, an additional objective for this study is to find meaningful features that can help explain different aspects of highly specific discussions. Many of the features described above, like N-grams or tf-idf, might have good predictive power but they are not easily interpretable and bear little relation to our codebook.

When considering NLP techniques applied to the educational domain, there is an increasing interest in developing models that capture important components of the construct to measure. Rahimi et al. \shortcite{Rahimi:17}, for example, developed a model for automated essay scoring using rubric-based features; Loukina et al. \shortcite{Loukina:15} evaluated different feature selection methods to obtain interpretable features in an educational setting.

In order to create an interpretable feature set we started by manually selected meaningful features from Speciteller (imageability, subjectivity, polarity, and familiarity ratings, number of connectives, fraction of stopwords). At training/test time, this set is combined with features from the \textit{Pronoun}, \textit{Named entities}, and \textit{Book} feature sets. Since all the features from the last 3 sets are interpretable, we only chose a few features from each set, selecting the ones with highest information gain with respect to specificity. For each fold, we first rank features in the \textit{Pronoun}, \textit{Named entities}, and \textit{Book} sets by information gain, then select the top k (based on the number of features in each respective set), concatenate them to the interpretable Speciteller features and train a logistic regression model. Section 5.4 will give examples of selected features.

\section{Experiments and Results}
In this section we provide results for our experiments. All classifiers and feature sets were evaluated using 10-fold cross validation, and using quadratic-weighted Cohen's kappa as the performance metric since it is important to make a distinction between different classification errors (e.g. classifying a low specificity turn at talk as high should result in bigger error than classifying it as medium).
We used the scikit-learn Python package\footnote{http://scikit-learn.org/stable/} for training and evaluating classifiers, as well as performing feature selection.
Specifically, sections 5.1 and 5.2 will be used to test our first hypothesis: that by retraining an existing model on our corpus we will obtain an improvement in performance.
Sections 5.2 and 5.3 will be used to test our second hypothesis: that by using features from additional NLP literature we can further improve the performance of a state-of-the-art model.
Section 5.4 will test our third hypothesis: that the additional features we handcrafted to capture specificity with respect to verbal discussion in an educational setting will lead to better performance.

\subsection{Baseline using Speciteller off-the-shelf}
\begin{figure}
  \includegraphics[width=\linewidth]{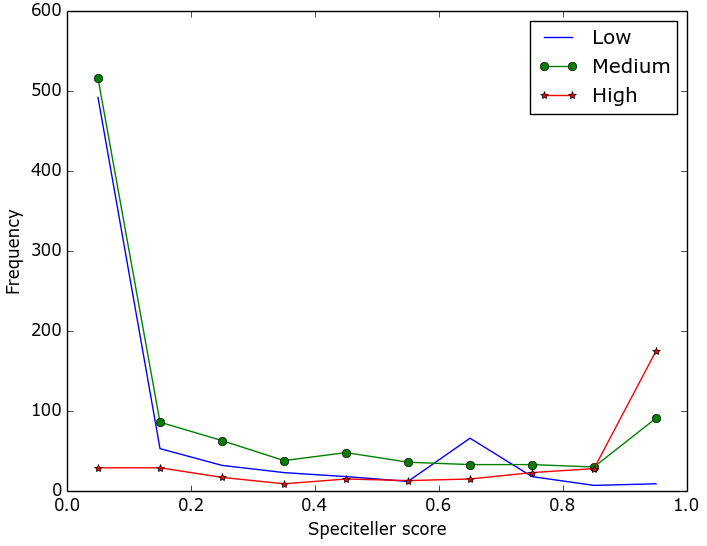}
  \caption{Speciteller scores by specificity class.}
  \label{fig:speciteller_distribution}
\end{figure}

Since we plan to use Speciteller as a baseline for comparing the performance of our proposed method, we iteratively tested thresholds to find the set which results in the highest quadratic-weighted kappa in all scenarios described in Section 4.1. The best result was obtained when the input to Speciteller is the complete turn at talk, and the resulting quadratic-weighted kappa is $0.495$, which represents Speciteller's upper bound performance.
Figure \ref{fig:speciteller_distribution} shows the frequency distribution of speciteller scores for each specificity class.

From the figure we can see that Speciteller is able to correctly capture specificity for a portion of the turns at talk in the dataset, as there is a peak in the low end of the spectrum for the distribution of low specificity scores and a peak in the high end of the spectrum for the distribution of high specificity scores. The medium specificity class seems to be the most problematic one, which has a similar trend as the low specificity class distribution in the low end of the spectrum, and a similar trend to the high specificity class distribution in the high end of the spectrum. Ideally we would expect the medium specificity distribution to have a peak towards the middle of the spectrum but that is not the case. Additionally, the low specificity class distribution shows a peak between $0.6$ and $0.7$ which will further penalize accuracy.

Table \ref{tab:confusion_matrix}
shows the confusion matrix when applying the optimal thresholds in order to get specificity labels from Speciteller scores.
\begin{table}[b]
\centering
\begin{tabular}{|c|l|c|c|c|}
\hline
                              &               & \multicolumn{3}{c|}{\textbf{predictions}}            \\ \hline
                              &               & \textbf{low} & \textbf{med} & \textbf{high} \\ \hline
\multirow{3}{*}{\textbf{ground truth}} & \textbf{low}  & 352          & 360          & 18             \\ \cline{2-5} 
                              & \textbf{med}  & 280          & 565          & 129            \\ \cline{2-5} 
                              & \textbf{high} & 4            & 139           & 210           \\ \hline
\end{tabular}
\caption{Confusion matrix using Speciteller scores to classify 
according to the optimal split points.}
\label{tab:confusion_matrix}
\end{table}
As we can see from the confusion matrix the overlap between the low and medium specificity classes and the medium and high specificity classes causes a large number of misclassifications: almost half of the low specificity turns at talk are classified as medium, over $40\%$ of the medium specificity turns at talk are classified as either low or high, and almost $40\%$ of high specificity turns at talk are classified as medium.
We believe these errors stem from two main reasons: as with many data-driven approaches, Speciteller is highly dependent on its training corpus. Speciteller was trained on articles from the Wall Street Journal and the New York Times. Articles written by professional writers are inherently different from transcriptions of spoken discussions between high school students.
Additionally, for training the model, Speciteller used a binary general/specific label, while we consider three labels in our work. Since Speciteller has no prior knowledge on medium specificity sentences, it is understandable that most of the misclassifications come from this class.

\subsection{Training using Speciteller features}
\label{sec:feature_sets_classification}
Our hypotheses as to why Speciteller does not work effectively out of the box  are related to its corpora and the way it was trained. With respect to the features used by Speciteller, we believe they might be useful in the context of classroom discussion as well. We extracted the shallow feature set and the neural network word embeddings feature sets and combined them to train a logistic regression classifier on our dataset. This classifier was chosen because one of our objectives is to compare the importance of other feature sets in addition to the \textit{Speciteller} one, and in order for this comparison to be fair we decided to use the same classifier Speciteller uses. Additionally, the classifier weights can be used to understand the importance of each feature. It is important to note that, unlike Speciteller, we will be using a single classifier on the combination of all features, and will not be able to leverage semi-supervised co-training.

Table \ref{tab:results1} shows the performance of a logistic regression classifier trained on this feature set and others described in the previous section. 
\begin{table}[h]
\centering
\begin{tabular}{|p{5cm}|p{1.7cm}|}
\hline
	\textbf{Feature sets} & \textbf{QWKappa}\\ \hline
	Speciteller & 0.5758\\ \hline 
	Speciteller + Online dialogue & 0.6347*\\ \hline 
	All: Speciteller + Online dialogue + Pronoun + NE + Book & 0.6360*\\ \hline 
    Speciteller + Semantic + Embeddings & \textbf{0.6550}* \\ \hline 
    Pedagogical & 0.5886 \\ \hline
\end{tabular}
\caption{Classification performance of different feature sets. * indicates statistically significant improvement over Speciteller features with p-value $<$ 0.001. Statistical significance was tested using a two-tailed paired t-test. Bold font highlights  best results.}
\label{tab:results1}
\end{table}
As we can see from the table, training a classifier using the Speciteller 
feature set on our corpus results in a considerable increase in performance, with
QWKappa of 0.5758 which represents a $16\%$ relative improvement over the 0.495 QWKappa obtained using Speciteller out of the box. 
This confirms our first hypothesis that Speciteller's performance, like many other methods, is highly dependent on its training corpus and using this model out of the box would give sub-optimal results.

\subsection{Speciteller and online dialogue features}
To test whether features from Section \ref{sec:literature_features} are  useful, we combined the Speciteller features with the Semantic, Lexical, and Syntactic features  and trained a logistic regression classifier based on the concatenated feature vectors. Table \ref{tab:results1} confirms our hypothesis that the 4 feature sets combined result in statistically significant (using a two-tailed paired t-test) higher kappa than using only Speciteller features. When combining Speciteller with each of the 3 other feature sets individually, kappa increases but  not with statistical significance.
We evaluated additional classifiers 
(Support Vector Machine, decision tree, random forest, Naive Bayes) but none of them outperformed logistic regression. Since the number of features is over 7000, we also tried using Recursive Feature Elimination (RFE) and Principal Component Analysis (PCA) for feature selection/reduction, but neither improved performance.

\subsection{Additional features}
To the feature set described in the previous section, we added the features described in Section 4.4. We then tested our third hypothesis by evaluating the performance of a logistic regression model trained with these features.

We can see from Table \ref{tab:results1} that all additional feature sets yield better 
performance than the $Speciteller$ feature set by itself. This result confirms our third 
hypothesis: the additional feature sets are able to capture aspects of specificity with respect to verbal discussion and the educational domain.
In particular the feature set containing neural network-based sentence embedding achieved the best kappa measure of $0.6550$, which suggests that sentence embeddings are also domain-dependent.
Compared to using Speciteller off-the-shelf this method improves kappa by $32\%$.
While the size of the neural network was constant during training/test (not optimized for each fold), we experimented with several numbers of hidden nodes (ranging from 50 to 200) for the LSTM and fully-connected layers, which resulted in kappa values in the range $0.6283 - 0.6550$.

The Pedagogical feature set is also able to marginally outperform the Speciteller feature set. Compared to the best result, the loss in kappa when using the Pedagogical set is $11\%$. At the expense of a slightly lower accuracy we gain the ability to use only informative features, which can be used to better understand  highly specific versus general classroom discussions.
The use of logistic regression also makes this possible: the model's coefficients give us an indication of how important features are. Table \ref{tab:coefficients} shows the top 12 features in the Pedagogical feature set ranked by the magnitude of the model's coefficients.

\begin{table}[t]
\centering
\begin{tabular}{|p{0.62\linewidth}|c|}
\hline
	\textbf{Feature} & \textbf{Coefficient} \\ \hline
	Number of connectives & 1.9168 \\ \hline
	\textit{Cosine similarity – whole summary} & 0.9293 \\ \hline
	MRC imageability & 0.8172 \\ \hline
    \textit{Number of characters} & 0.6931\\ \hline
	MPQ subjectivity & -0.5440 \\ \hline
	Fraction of stopwords & -0.4087 \\ \hline
    MRC familiarity & 0.3986 \\ \hline
    \textit{Number of possessive pronouns} & 0.2035 \\ \hline
	\textit{Number of named entities normalized} & 0.1865 \\ \hline
    \textit{Number of 3\textsuperscript{rd} person pronouns} & 0.1755 \\ \hline
	\textit{Word overlap – whole summary} & 0.1585 \\ \hline
    \textit{Number of personal pronouns} & 0.1476 \\ \hline
\end{tabular}
\caption{Pedagogical feature set and respective logistic regression coefficients. Italic font  shows features developed in this study (Section 4.4).}
\label{tab:coefficients}
\end{table}
The table shows the results of a model trained on the complete dataset.
The number of connectives seems to be the most important feature for predicting 
high specificity. This seems straightforward, as more connectives translates into more clauses, which provide more information. While the annotators did not look for connectives during coding, one of the aspects they analyzed was the presence/absence of a chain of reasoning, and the number of connectives might capture that aspect.
The cosine similarity between the turn at talk and the book summary (considered as one entity) is another important feature in the model: higher similarity between the summary and what a student says means that they are using terms from the book. This feature seems to capture another aspect in our codebook, the use of book-specific vocabulary.
We can use the information provided by these features to understand specificity, and to give feedback to teachers and students: if for example a student tends to produce low specificity turns at talk and the number of connectives used is generally low, that might be an indication that they should elaborate more on their statements.
Conversely, if the number of connectives used is high but the number of characters mentioned is low, that might be an indication that the student should reference specific characters more often.

\section{Conclusions and Future Work}

We proposed several models for predicting specificity and evaluated them on   text-based, high school  classroom discussion data. We showed that an existing general-purpose system  achieves significantly better performance when its features are used for retraining on educational data. We also showed that performance can be further improved by using additional features from the NLP literature  \cite{Swanson:15}, especially when combined with neural network embeddings and other new features tailored to text-based classroom discussion. Finally we proposed a subset of pedagogical features which, even though slightly less performing, provide the ability to interpret the features, which is especially important for the educational community.

As more data becomes available, we will explore more advanced neural network models  and examine method generalization (e.g., social science vs. ELA,  middle vs. high school).  
We also plan to 
analyze features at a finer granularity than a turn at talk
and to extract the book summary features automatically from the original texts.
Since our dataset is already annotated for argument type, and will be annotated for  discussion quality, we plan to investigate  relationships between specificity, argumentation, and quality. 

\section*{Acknowledgements}
We want to thank Dr. Amanda Godley, Christopher Olshefski, Zane Denmon, Zinan Zhuang, Clare Miller, Keya Bartolomeo, and Annika Swallen for their contribution, and all the anonymous reviewers for their helpful suggestions.

This work was supported by the Learning Research and Development Center at the University of Pittsburgh.

\bibliography{emnlp2017}
\bibliographystyle{emnlp_natbib}

\end{document}